# LILA-BOTI : Leveraging Isolated Letter Accumulations By Ordering Teacher Insights for Bangla Handwriting Recognition

Md. Ismail Hossain*, Mohammed Rakib*, Sabbir Mollah*, Fuad Rahman† and Nabeel Mohammed*

*Apurba-NSU R&D Lab, Department of Electrical and Computer Engineering

North South University, Dhaka, Bangladesh

†Apurba Technologies

440 N. Wolfe Rd., Sunnyvale, CA 94085, USA

Email: *{ismail.hossain2018, mohammed.rakib, sabbir.mollah, nabeel.mohammed}@northsouth.edu

†{fuad}@apurbatech.com

*Abstract*—Word-level handwritten optical character recognition (OCR) remains a challenge for morphologically rich languages like Bangla. The complexity arises from the existence of a large number of alphabets, the presence of several diacritic forms, and the appearance of complex conjuncts. The difficulty is exacerbated by the fact that some graphemes occur infrequently but remain indispensable, so addressing the class imbalance is required for satisfactory results. This paper addresses this issue by introducing two knowledge distillation methods: Leveraging Isolated Letter Accumulations By Ordering Teacher Insights (LILA-BOTI) and Super Teacher LILA-BOTI. In both cases, a Convolutional Recurrent Neural Network (CRNN) student model is trained with the dark knowledge gained from a printed isolated character recognition teacher model. We conducted inter-dataset testing on *BN-HTRd* and *BanglaWriting* as our evaluation protocol, thus setting up a challenging problem where the results would better reflect the performance on unseen data. Our evaluations achieved up to a 3.5% increase in the F1-Macro score for the minor classes and up to 4.5% increase in our overall word recognition rate when compared with the base model (No KD) and conventional KD.

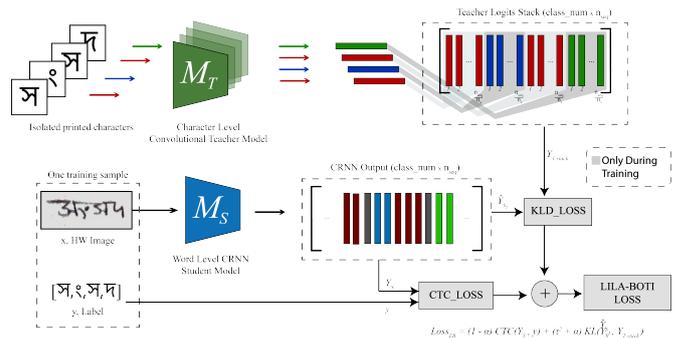

Figure 1. The aim of the LILA-BOTI approach is to take insights from the printed isolated characters through a teacher model $M_T$ and use those insights to train a student CRNN model $M_S$. This figure illustrates how the LILA-BOTI loss is calculated during the training phase. For a sample word x, the isolated character images representing the graphemes present in x's label are selected from the printed isolated character dataset $D_C$. These images are fed to $M_T$ to get the teacher outputs. The outputs are copied and concatenated according to our stacking algorithm to form a stack of logits. This stack is then used to calculate its KL Divergence distance with the CRNN output.

## I. Introduction

Handwritten word recognition models have an immense value to businesses and governments, involving finance, law, and creativity sections that have accumulated an abundance of handwritten documents. The field of Handwritten Text Recognition (HTR) gained more popularity with the advent of deep learning and the progress in computation power. Modern approaches allow satisfactory results in high-resource languages such as English. However, it is challenging to gain acceptable results for languages that suffer from a scarcity of extensive datasets. To complicate the whole process, character distribution in any natural language is found to be imbalanced approximately following Zipf's law [1]. This problem is more evident when dealing with Bangla alphabets due to the language's morphological richness.

The Bangla alphabet is mainly divided into consonants and vowels. These alphabets can then form conjuncts and diacritics that are visually different from their base forms. The frequency of the usage of these large numbers of graphemes varies significantly. Training a CRNN model with an imbalanced dataset leads the model to perform well on words containing frequently used characters while poorly on words with less frequent characters. This paper proposes a new way of using knowledge distillation in Connectionist Temporal Classification (CTC) loss-based CRNN model training to address these data imbalance issues. Our aim is to transfer the insights gained from the teacher model to the CRNN student model.

The teacher model can be any model capable of classifying an isolated character image. In our proposed method, this model is trained on a synthetically generated balanced printed isolated character dataset. The character-level teacher model was chosen because it is easy to generate a balanced character-level dataset while getting a natural word-level dataset that has balanced characters is almost impossible. The student model is a CRNN model trained on an inherently imbalanced handwritten word-level dataset. The intuition is that dark knowledge gained from the teacher model, trained on a balanced dataset, will impede the student model from getting biased towards the major classes of the imbalanced word-level dataset. The selection of this approach poses two fundamental challenges. The fact that teacher model is trained on printed data, while our target is to recognize handwritten data. The second point is that the teacher is a character-level model while the student is a word-level model that predicts a sequence of characters. The first challenge is trivially addressed by the loss function structure in the knowledge distillation approach, where a parameter regularizes the terms. We take on the second challenge by proposing a stacking algorithm to match the teacher model's output with the student model's one to allow knowledge distillation. We refer to this approach as Leveraging Isolated Letter Accumulations By Ordering Teacher Insights (LILA-BOTI). Fig. 1 depicts the training pipeline of LILA-BOTI in detail for a single training sample. We improve upon this approach with further observations and acquire better results by proposing a Super Teacher LILA-BOTI approach where the teacher model is trained on a superset of graphemes, including graphemes not present in the student model training dataset.

The main contributions of this paper are summarized as follows:

- We introduce LILA-BOTI and Super Teacher LILA-BOTI, two new knowledge distillation methods for the CRNN model. In these approaches, the student model is trained on an imbalanced word-level dataset by leveraging the teacher insights (dark knowledge) from a model trained on an isolated printed characters dataset.
- To evaluate the efficacy of our proposed methods, we have compared our knowledge distillation approach with a base CRNN model and a conventional knowledge distillation approach. We demonstrate that our approaches can give better results on both *BanglaWriting* and *BN-HTRd* datasets, especially in the minor classes.
- To the best of our knowledge, we are the first to propose and utilize an inter-dataset evaluation protocol for Bangla word-level handwriting recognition models to obtain results that evaluate the performance of our proposed methods rigorously. This evaluation framework adds more weight to the metrics as it is now arduous for the models to get better results.
- Our methods improve the F1-Macro scores on the minor classes of *BN-HTRd* by 1.1% when evaluated on BanglaWriting and the minor classes of *BanglaWriting* by 3.5% when evaluated on *BN-HTRd* compared to the base model (No KD) and conventional KD approach. Besides, our techniques enhance the word recognition rate of the student model by 1.3% for *BN-HTRd* when trained on *BanglaWriting* and by 4.5% for *BanglaWriting* when trained on *BN-HTRd*.

Code for this work is available at https://github.com/Apurba-NSU-RnD-Lab/LILA-BOTI

## II. Related Works

The concept of knowledge distillation was first introduced by Caruana et al. [2] and later formulated by Hinton et al. [3]. Caruana et al. [2] showed that knowledge from a large ensemble model could be compressed to a small model by training the small model with the logits generated by the large ensemble model. Later on, Hinton et al. [3] generalized the concept with appropriate formulas and experiments on MNIST and Voice Recognition problems. They used probabilities generated by the softmax activation function instead of logits. Their paper describes the softened softmax activations as the *dark knowledge*, which we consider as teacher insights. In this paper, we use a modified version of the knowledge distillation loss function deduced by Hinton et al. [3] which is broadly described in the Experimental Setup section. Aside from using knowledge distillation for model compression, Sarfraz et al. [4] shows that it can be effectively used for other problems such as learning label noise and class imbalance as well. In our experiments, we utilize the KD framework for leveraging insights from a teacher model trained on a balanced character-level dataset.

The introduction of Connectionist Temporal Classification [5] made it possible to create an end-to-end unsegmented sequence labeling model. Shi et al. [6] uses the CTC loss to train an end-to-end model for image-based text sequence recognition. In Bangla OCR literature, several works have been done on isolated character recognition [7]–[11]. Omee et al. [8] elaborately discusses the workflow and algorithms to develop Bangla OCR. Basu et al. [9] developed a Multi-Layer Perceptron (MLP) based classifier for Bangla handwritten character recognition utilizing a 76 element feature set. They report an accuracy of 86.46% on the training set and 75.05% on the test set of their manually created handwritten dataset. The authors of [10] also tried to develop an improved Bangla OCR with their own custom dataset. Shopon et al. [11] leverages unsupervised pretraining using autoencoder with Deep ConvNet to improve the performance of Bangla handwritten digit recognition. As we see, a lot of work has been done to improve Bangla OCR, but none of them particularly addressed the performance of minor classes. In this paper, we emphasize the performance of the minor classes and try to improve it using knowledge distillation.

## III. Datasets

### A. Synthetically Generated Isolated Printed Characters

Our teacher model is an isolated character prediction model. The training dataset for the teacher model is generated synthetically by using the Text Recognition Data Generator module [12] from Python Package Index. A list of various Bangla fonts was fed to the package, along with a set of graphemes. For each grapheme in the grapheme set, arbitrarily 160 isolated character images were generated using random augmentations.

*B. Handwritten Word Datasets*

In our experiments, we have used two handwritten word-level datasets to train our student model. The *BanglaWriting* [13] dataset has 21,234 Bangla handwritten words. Additionally, the *BN-HTRd* dataset [14] has 108,147 Bangla handwritten words. For our inter-dataset evaluation protocol, we have picked one of these two datasets for training, and the other for testing and then repeated the process by interchanging the roles of the datasets.

*C. Grapheme Extraction*

Figure 2. Two examples are shown where the same grapheme can have multiple Unicode representations.

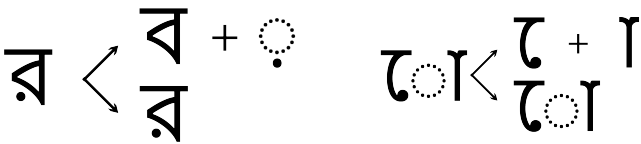

Bangla alphabet can be divided into consonants, consonant conjuncts, vowels, vowel diacritics, numbers, marks, and symbols. Each of these subdivisions may consist of a large inventory of graphemes. Additionally, some graphemes can have multiple Unicode representations, as shown in Fig. 2. Hence, it is necessary to normalize the word labels to make sure each occurrence of a grapheme is extracted consistently. In our experiments, we regard consonants with diacritics and consonant clusters as single graphemes. Following these preprocessing, we find 213 graphemes in the *BN-HTRd* dataset and 177 graphemes in the *banglawriting* dataset. The superset containing graphemes from both sets has a cardinality of 219. And the intersection has a cardinality of 171. *BN-HTRd* has 42 unique graphemes and *BanglaWriting* has 6 unique graphemes.

*D. Data Imbalance*

Bangla can have a large number of character classes independent of the grapheme extraction methods. Hence, it is expected to have an imbalance on a grapheme level. In fact, we can notice the longtail feature of the grapheme counts of *BN-HTRd* and *BanglaWriting* in Fig 3 which denotes in log scale the number of graphemes for a given character class. For our work, we have defined minor classes based on the support count of each grapheme/class, and it will vary dataset-wise. All the classes whose support count is less than 10% of the most frequently occurring class are considered to be minor classes

Figure 3. The grapheme count for both datasets *BN-HTRd* (Maroon) and *BanglaWriting* (Blue) are illustrated along the y axis which follows a base 10 log scale. The grapheme classes for each dataset align along the x axis. For example, along the x axis the first *BN-HTRd* class indicates the same first *BanglaWriting* class. The higher number of total graphemes available in both datasets appears on the left along the x axis. The constantly decreasing slope in this graph indicates an exponential decrease due to the logarithmic scale, which points toward a significant grapheme-level data imbalance.

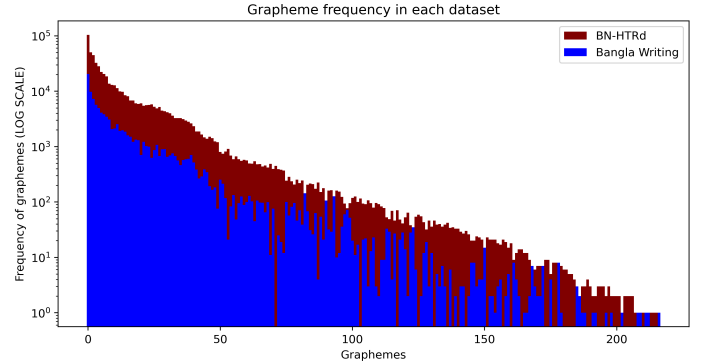

### IV. Proposed Methodology

Given a handwritten word image x consisting of $n_x$ graphemes from an imbalanced word level dataset $D_S$, our aim is to train a CRNN model $M_S$ with $D_S$ such that its parameters are regularized with respect to a teacher model $M_T$ trained on a balanced printed isolated character dataset $D_T$. The set of graphemes that form the words in $D_S$ is, $G_S = \{ g_S^1, g_S^2, ..., g_S^p \}$ Also, let the set of graphemes used to generate $D_T$ be $G_T = \{ g_T^1, g_T^2, ..., g_T^p \}$. $G_T$ is generated in the method specified in section III-A.

*A. Leveraging Isolated Letter Accumulations By Ordering Teacher Insights (LILA-BOTI)*

Here we describe our proposed method LILA-BOTI. It is a knowledge distillation approach in which a word-level student model learns from the insights extracted from a balanced printed character dataset through a character-level teacher model. In this case, the teacher model is trained on dataset $D_C$, which is generated on grapheme set $G_T = G_S$. The main purpose of the teacher model is to provide insights on the set of graphemes present in the training set of the student model, albeit that the insights will come from printed isolated characters.

For each grapheme available in a training word sample, we choose an equivalent character image from $D_T$. These character images are then inferred through the teacher model to get their respective scores. While getting the scores, it is ensured that the scores match the labels. If a particular score doesn't match its corresponding label, a new character image is taken from $D_T$ until the score matches the label. Next, these scores are copied and stacked together into a stack of a size equivalent to the student's output. This stacking approach is detailed in the Algorithm 1. The student model returns a sequence $n_{seq}$ score vectors each of $|G_S|$ dimensions. The stacking algorithm also provides us with a sequence of scores that matches the shape of the student's output. These can now be used to derive soft probabilities. To soften probabilities, we use a hyperparameter called temperature ($\tau$). When $\tau = 1$, we get the normal output from softmax. But as we increase $\tau$, the softmax output becomes softer, indicating which classes our teacher model found to be more similar to the predicted class. The authors of [3] call this dark knowledge that is embedded within the teacher model. And this dark knowledge is transferred to the student model via the distillation process. It is to be noted that we use the same value of $\tau$ while softening the teacher's targets and student's logits as done in [3].

**Algorithm 1** The stacking process.
1: **procedure** PREDICTIONSSTACKING(LABEL)
2: $\quad$ *graphemes ← extractGraphemes(label)*
3: $\quad$ $n_x$ ← size of graphemes
4: $\quad$ *stack ← ()*
5: $\quad$ **for** i ← 0 to $n_x$ **do**
6: $\quad\quad$ *printedCh ← selection(graphemes[i])*
7: $\quad\quad$ *out ← $M_T$(printedChar)*
8: $\quad\quad$ **for** j ← 0 to floor($\frac{n_{seq}}{n_x}$) **do**
9: $\quad\quad\quad$ *stack.append(out)*
10: $\quad$ Fill *stack*[floor($\frac{n_{seq}}{n_x}$)+1 to $n_x$] with last *out*
11: $\quad$ **return** *stack*.

Let (x,Y) in $D_S$ where x is a handwritten word image and Y is the corresponding sequence of $n_x$ one-hot vectors as label. Now our student model $M_S$, given an input x will output logits $L_S$ which can be shown as $L_S = M_S(x)$. These logit values are softened using the temperature ($\tau$) and used in the softmax function $\sigma$ to get the soft probabilities denoted by $\hat{Y}_{S\tau}$ in $\hat{Y}_{S\tau} = \sigma(L_S/\tau)$. On the other hand, $Y_S$ denotes the hard probabilities in $Y_S = (y_S^1, ..., y_S^{n_{seq}}) = \sigma(L_S)$ to be used by the CTC loss.

Now, let $G_y$ be the sequence of the $n_x$ numbers of graphemes extracted from Y. And $I_y$ is the sequence of isolated printed character images from $D_C$, which align with the elements of $G_Y$.

The teacher model $M_T$, will extract the character prediction scores of each image in $I_y$ and these scores will be stacked together to form a sequence of predicted scores. So for each character image i in $I_y$ the logits can be denoted as $L_{T_i} = M_T(I_{y_i})$ and the probability distribution for each character can be shown as $Y_{T_i} = (y_{T_i}^1, ..., y_{T_i}^{n_{seq}}) = \sigma(L_{T_i}/\tau)$. The stack of character distributions for a single word then can shown as:

$$Y_{T-stack} = \text{CONCAT}_{i=0}^{n_x} \text{CONCAT}_{j=0}^{n_{seq}/n_x}(Y_{T_i}) \quad (1)$$

The LILA-BOTI Loss Function, $Loss_{LB}$ can then be expressed as:

$$Loss_{LB} = (1-\alpha)CTC(Y_S, y) + (\tau^2 + \alpha)KL(\hat{Y}_{S\tau}, Y_{T-stack}) \quad (2)$$

Since the LILA-BOTI loss calculation is done only during model training, the teacher model is not required during model inference. This implies that the CRNN model trained with LILA-BOTI loss will not have any added time complexities with respect to an equivalent CRNN model. During inference, given a word image x, $M_S$ will simply output a sequence of probabilities without any extra overhead.

*B. Super Teacher LILA-BOTI*

In our paper, we are using knowledge distillation to transfer the dark knowledge embedded within the teacher model to the student model. Hinton et al. [15] has shown through experiments that after excluding all samples of a class from the training dataset, it is still possible to make the model recognize samples of the excluded class through knowledge distillation. From that finding, we hypothesize that a teacher trained on a more diverse set of graphemes can potentially provide more informative dark knowledge which may help the student model on recognizing the minor classes.

Following this intuition, we propose Super Teacher LILA-BOTI with a superset of graphemes $G_T$. In the previously explained LILA-BOTI approach, the teacher model is trained on the grapheme set $G_T = G_S$. We now propose a scenario where the $G_T$ is not necessarily equal to $G_S$ rather, is a superset of $G_S$. In theory, $G_T$ can have all the possible graphemes in the Bangla language. However, in our implementation, we have created this superset by taking the union of the graphemes in *BN-HTRd* and *BanglaWriting*. This does not break our inter-dataset evaluation protocol since the teacher model gets trained on synthetic data, and no image from the word-level test set is directly contributing to the training. In our result section, we demonstrate that the Super Teacher LILA-BOTI can obtain better and more consistent results compared to the LILA-BOTI approach.

## V. EXPERIMENTAL SETUP

*A. Model Architecture*

*1) Teacher:* In this work, we implement two types of teacher models - a shallow conv2 [16] model, and a deeper ResNet18 [17] model. Throughout the paper, we define the teacher model as $M_T$, and it is trained with the synthetically generated isolated printed character dataset $D_T$ containing images of graphemes from $G_T$. The output logits of $M_T$ are scores over graphemes in $G_T$. The difference in the performance of our student model trained by using insights from conv2 and ResNet18 is discussed in section VI (Results).

*2) Student:* $M_S$ is implemented as a CRNN model with a VGG backbone and a BiLSTM layer [18]. The output of the student model will be a sequence of $n_{seq}$ vectors, where each vector contains predictive scores over the graphemes in $G_S$. In this work, $n_{seq}$ is set to 31. This number is chosen based on the maximum grapheme length of the words, and it is kept constant to make sure it is possible to generate sequences of the same lengths from both the teacher model and the student model. Having $n_{seq}$ as a constant value also allows the model to be trained on a GPU, which requires fixed sized data batches.

Table I
QUANTITATIVE RESULTS FROM THE INTER-DATASET EVALUATION PROTOCOL.

| Train Set | Test Set | KD Type | NED | CRR | WRR | F1 Macro | | |
|---|---|---|---|---|---|---|---|---|
| | | | | | | All Class | Minority Class | Majority Class |
| BanglaWriting | BN-HTRd | No KD | 26195 | 73.55 | 48.02 | 35.17 | 25.25 | 70.30 |
| | | Conventional KD | 30758 | 68.51 | 41.63 | 30.10 | 20.29 | 64.81 |
| | | LILA-BOTI Teacher: resnet18 | 25465 | 74.08 | **49.38** | 37.42 | 27.76 | **71.61** |
| | | Super Teacher LILA-BOTI Teacher: resnet18 | 26071 | 73.74 | 48.14 | 35.35 | 25.38 | 70.65 |
| | | LILA-BOTI Teacher: conv2 | 26286 | 73.92 | 48.66 | 37.83 | 28.19 | 71.69 |
| | | Super Teacher LILA-BOTI Teacher: conv2 | 25503 | **74.40** | 49.31 | **38.20** | **28.82** | 71.42 |
| BN-HTRd | BanglaWriting | No KD | 4302 | 76.67 | 51.72 | 45.33 | 39.67 | 76.07 |
| | | Conventional KD | 3558 | 82.16 | 58.49 | 55.77 | 50.18 | **82.40** |
| | | LILA-BOTI Teacher: resnet18 | 3349 | 82.82 | 59.09 | 54.93 | 50.00 | 81.86 |
| | | Super Teacher LILA-BOTI Teacher: resnet18 | 3063 | **84.46** | **63.22** | **56.09** | **51.32** | 82.13 |
| | | LILA-BOTI Teacher: conv2 | 3353 | 82.74 | 58.66 | 55.34 | 50.55 | 81.50 |
| | | Super Teacher LILA-BOTI Teacher: conv2 | 3178 | 83.58 | 61.79 | 54.63 | 49.73 | 81.20 |

### B. Comparison Baseline

A literature survey of previous works did not yield any paper that reported results using the *BN-HTRd* dataset. The few papers that did use *BanglaWriting* [19] [20] dataset, did not clarify their training set, validation set and testing set partitions. Meaning an intra dataset comparison with the previous work using these two datasets are not possible. For ease of comparisons in future works, we are releasing our partitions in the GitHub repository. To estimate the integrity of the LILA-BOTI approaches, we used a base CRNN model with CTC loss and a CRNN model trained with CTC loss along with a conventional knowledge distillation approach as our baselines. The Conventional Knowledge Distillation approach uses the same CRNN model previously explained as both the teacher and the student model. The main intuition behind this approach is that, since we are using the same model architecture for both the teacher and the student models the sequence of their predictions align with each other, therefore there is no need for an extra stacking algorithm like in LILA-BOTI approaches. If $M_{CT}$ is the conventional teacher CRNN model and $M_S$ is the student CRNN model, then given a sample training (x,Y), the loss, $Loss_{CL}$ for Conventional KD can be given as:

$$Loss_{CL} = (1-\alpha)CTC(Y_S, Y) + (\tau^2 + \alpha)KL(Y_S, Y_{CT}) \quad (3)$$

In (3), the logits, $L_T$ from the teacher model $M_{CT}$ is denoted as $L_T = M_{CT}(x)$ and the softened probability distribution from the conventional teacher model, $Y_{CT}$ is depicted as $Y_{CT} = \sigma(L_T/\tau)$. $Y_S$ is the hard probabilities from the student model. As we observe, the change is only in the KL-Divergence part of this equation, where both inputs are from two CRNN models. These two terms already have the same size. Hence it doesn't need the stacking algorithm required for LILA-BOTI.

## VI. RESULTS

### A. Experiment Results

As shown in Table I the results are separated in two groups. In the first group, the approaches are trained on *BN-HTRd* dataset and evaluated on *BanglaWriting* dataset. In the second group, the training and testing datasets were flipped. This inter-dataset evaluation protocol gives more rigorous results on the performance of unseen data. In each group, we conducted six experiments.

The No KD approach indicates the base CRNN model without any Knowledge Distillation. This model was trained to set a baseline for our evaluations. The base model achieves 73.55% and 76.67% character recognition rates when trained with *BanglaWriting* and *BN-HTRd* respectively. We notice a significant discrepancy between the major and minor class F1-Macro scores. In fact, in the first group, the difference between the major and minor classes F-1 Scores is 45.05%. Reducing this discrepancy would give a better CRR and WRR. This is a problem that this paper will try to address.

Results indicate that the conventional KD approach performs poorly across all metrics in the first evaluation group and improves performance across all metrics in the second evaluation group when compared with the base model results. This may be a reflection of the fact that for the first evaluation group, the train set, i.e., *BanglaWriting* has significantly less no. of samples (21,234 words) with less data diversity (5,470 unique words). Hence, there might be an element of overfitting which causes the teacher model to provide less useful insights. However, for the second evaluation group, the train set, i.e., *BN-HTRd* has a large no. of samples (108,147) with large data diversity (23,115 unique words). As a result, the teacher model may provide more effective insights.

LILA-BOTI with a deeper teacher model (ResNet18) and LILA-BOTI with a shallow Conv2 model outperform No KD and Conventional KD across all metrics, including minor classes for the first training group. This shows that in LILA-BOTI, the teacher insights are helping the student model learn better the features of minor classes. However, for the second evaluation group, we observe that the performance of LILA-BOTI is neck and neck with Conventional KD.

To improve the performance in inter-dataset evaluation settings for the LILA-BOTI approach, Super Teacher LILA-BOTI experiments were conducted. Instead of training the teacher model with only the grapheme occurrences in the training dataset, the superset of the graphemes of both datasets was used in this setup. This approach consistently improved upon the results of the Base Model (No KD), Conventional KD, and LILA-BOTI. It manages to increase CRR by 0.85% for the first evaluation group when compared with the base model and improves 2.3% for the second evaluation group when compared with the conventional KD. In the first group, LILA-BOTI improves the F1-Macro of minor classes by 3.57% with respect to NO KD. In the second group, Super Teacher LILA-BOTI improves the F1-Macro of minor classes by 1.14% with respect to Conventional KD. This is a more consistent increase across the datasets compared to the LILA-BOTI results.

*B. Comparison of Sample Output of Different Models*

We now present table II containing a few samples from our test sets highlighting some strengths and weaknesses of the various models used in our experiments. For the first sample word "ধরণের", the minor classes are "ধ" and "ণ." Observing Table II we see that all the models could correctly predict the minor class "ধ." However, only LILA-BOTI and Super Teacher LILA-BOTI could correctly predict the minor class "ণ". The next word "বর্তমান" consists of major classes only, and all the models could correctly predict all of the graphemes. For the third sample word "প্রভৃতি", the minor classes are "ভ" and "ৃ". None of the models could correctly predict the minor class "ৃ" properly. However, only LILA-BOTI and Super Teacher LILA-BOTI could correctly predict the minor class "ভ". For the last word "যৌথপরিবারে", the minor classes are "য", "ৌ", and "থ", and the rest are major classes. Observing the output of all the models, we see that none of them could correctly predict the entire word correctly with misprediction in both major and minor classes. LILA-BOTI and Super Teacher LILA-BOTI did not do as well as the earlier examples here. A possible reason for this is that our teacher model consists of printed characters. Hence, it is expected that the student model will be able to improve the results of the characters that most resemble printed fonts. In the last example, the handwriting is very personalized and does not resemble printed characters. Hence, none of the knowledge distillation approaches could improve the result of the last sample.

Table II
Comparison of Sample Input Output of Different Models

| Train Set | Test Set | Input From Test Set | KD Type | Label | Prediction | CRR (%) |
|---|---|---|---|---|---|---|
| BW | BN-HTRd | ধরণের | No KD | ধরণের | ধরনের | 80 |
| | | | Conventional KD | | ধরনের | 80 |
| | | | LILA-BOTI | | ধরণের | 100 |
| | | | Super Teacher LILA-BOTI | | ধরণের | 100 |
| | | বর্তমান | No KD | বর্তমান | বর্তমান | 100 |
| | | | Conventional KD | | বর্তমান | 100 |
| | | | LILA-BOTI | | বর্তমান | 100 |
| | | | Super Teacher LILA-BOTI | | বর্তমান | 100 |
| BN-HTRd | BW | প্রভৃতি | No KD | প্রভৃতি | গুরুটি | 0 |
| | | | Conventional KD | | প্রভৃতি | 50 |
| | | | LILA-BOTI | | প্রভৃতি | 85.71 |
| | | | Super Teacher LILA-BOTI | | প্রভৃতি | 66.67 |
| | | যৌথপরিবারে | No KD | যৌথপরিবারে | সোষভারিকো | 10 |
| | | | Conventional KD | | ঘোরিকারে | 11.11 |
| | | | LILA-BOTI | | যৌথাভরিকারে | 20 |
| | | | Super Teacher LILA-BOTI | | যৌথরিবাবে | 60 |

VII. Conclusion

Training a handwritten word recognition model for morphologically rich languages like Bangla is difficult due to many reasons, prominent amongst which is the natural imbalance in the graphemes of available word-level datasets. To address this issue, we proposed LILA-BOTI, a knowledge distillation approach that incites the word-level student model to learn from the insights gained from a teacher model trained on printed isolated characters. The proposed LILA-BOTI approach gained better results in both character recognition and word recognition rates; however, the noticeable increase was in the performance in the macro F1-Score for the minor classes. We also proposed an inter-dataset evaluation protocol using the *BanglaWriting* and *BN-HTRd* datasets, where models trained on one dataset are evaluated on the other. While LILA-BOTI reduced the performance gap between major and minor classes, our second approach, that of using a superset of graphemes in Super Teacher LILA-BOTI, achieved more consistent scores across all the metrics. This method improved the F1-Macro of the minor classes by 8.5% when evaluated on *BN-HTRd* and by 1.1% when evaluated on *BanglaWriting* with respect to conventional KD. This method also improves the WRR by 7.6% when evaluated on *BN-HTRd* and by 4.7% when evaluated on *BanglaWriting* with respect to conventional KD. We aim to experiment with our approach on other morphologically rich languages in the future.

Acknowledgement

The authors would like to acknowledge the encouragement and funding from the "Enhancement of Bangla Language in ICT through Research & Development (EBLICT)" project under the Ministry of ICT, the Government of Bangladesh.